\documentclass[11pt,a4paper]{article}
\usepackage{emnlp2020}
\usepackage{times}
\usepackage{latexsym}

\usepackage{url}
\usepackage{breakurl}

\usepackage{graphicx}
\usepackage[export]{adjustbox}
\usepackage{bm}
\usepackage{here}
\usepackage{amsmath}
\usepackage{amssymb}
\usepackage{amsfonts}
\usepackage{multirow}
\usepackage{arydshln}
\usepackage{enumitem}
\usepackage{dsfont}

\usepackage{algorithm}
\usepackage{algpseudocode}
\usepackage{soul}
\usepackage{float}
\usepackage{subfigure}
\usepackage{microtype}

\title{Few-Shot Intent Detection via Contrastive Pre-Training and Fine-Tuning}

\author{
  \textbf{Jian-Guo Zhang}$^1$\thanks{\ \ Work done while the first author was an intern at Adobe Research.}, \textbf{Trung Bui}$^2$, \textbf{Seunghyun Yoon}$^2$,   \textbf{Xiang Chen}$^2$, \textbf{Zhiwei Liu}$^1$ \\ \textbf{Congying Xia}$^1$, \textbf{Quan Hung Tran}$^2$, \textbf{Walter Chang}$^2$, \textbf{Philip Yu}$^1$ \\
$^1$ University of Illinois at Chicago, Chicago, USA \\
$^2$Adobe Research, San Jose, USA\\
  { \texttt{\{jzhan51, zliu213, cxia8, psyu\}@uic.edu},} \\ {\texttt{\{bui, syoon, xiangche, qtran, wachang\}@adobe.com}} \\
}
\date{}

\begin{document}
\maketitle
\begin{abstract}
In this work, we focus on a more challenging few-shot intent detection scenario where many intents are fine-grained and semantically similar. We present a simple yet effective few-shot intent detection schema via contrastive pre-training and fine-tuning. Specifically, we first conduct self-supervised contrastive pre-training on collected intent datasets, which implicitly learns to discriminate semantically similar utterances without using any labels. We then perform few-shot intent detection together with supervised contrastive learning, which explicitly pulls utterances from the same intent closer and pushes utterances across different intents farther.  Experimental results show that our proposed method achieves state-of-the-art performance on three challenging intent detection datasets under 5-shot and 10-shot settings. 
\end{abstract}

\section{Introduction}

Intent detection, aiming to identify intents from user utterances, is a key component in task-oriented dialog systems. In real systems such as Amazon Alexa, correctly identifying user intents is crucial for downstream tasks~\citep{zhang2020find,ham2020end}.  A practical challenge is data scarcity as it is expensive to annotate enough examples for emerging intents, and how to accurately identify intents in few-shot learning has raised attention.

Existing methods address the few-shot intent detection tasks mainly from two perspectives: (1) data augmentation and (2) task-adaptive training with pre-trained models. For the first category, \newcite{zhang2020discriminative} and \newcite{mehri2020example}  propose a nearest neighbor classification schema with full use of the limited training examples in both training and inference stages. 
\newcite{xia2020cg} and \newcite{peng2020data} propose to generate utterances for emerging intents based on variational autoencoder~\citep{kingma2013auto} and GPT-2~\citep{radford2019language}, respectively. 
For the second category,  \newcite{efficient_intent}  and \newcite{mehri2020dialoglue} conduct intent detection by leveraging related conversational pre-training models based on a few hundred million conversations.  Meanwhile, they devise a task-adaptive training schema where the model is pre-trained on all relative intent datasets or the target intent datasets with mask language modeling.

However, previous methods such as data augmentation related models~\cite{liu2021augmenting} are inefficient for training and hard to scale to tasks with lots of intents. Moreover, these models do not tackle well the following scenarios: In real scenarios, the few-shot intent detection could be more challenging when there exist many fine-grained intents, especially semantically similar intents. For instance, BANKING77~\citep{efficient_intent} has a single domain with 77 intents, and CLINC150~\citep{oos-intent} has ten domains with 150 intents. 
Many intents in the datasets are similar. Therefore, training models is rather challenging when there are only limited examples. 

Inspired by the recent success of contrastive learning~\citep{he2020momentum,gunel2020supervised,chen2020simple,radford2learning,liu2021fast,gao2021simcse,liu2021contrastive}, which aims to enhance 
discrimination abilities of models,
this work proposes improving few-shot intent detection via Contrastive Pre-training and Fine-Tuning (CPFT). Intuitively, we first learn to implicitly discriminate semantically similar utterances via contrastive self-supervised pre-training on intent datasets without using any intent labels. 
We then jointly perform few-shot intent detection and supervised contrastive learning. The supervised contrastive learning helps the model explicitly learn to pull utterances from the same intent close and push utterances across different intents apart. 


Our contributions are summarized as follows: 1) We design a simple yet effective few-shot intent detection schema via contrastive pre-training and fine-tuning. 2) Experimental results verify the state-of-the-art performance of CPFT on three challenging datasets under 5-shot and 10-shot settings.
\section{Related Work}
Since this work is related to few-shot intent detection and contrastive learning, we review recent work from both areas in this section. 

The few-shot intent detection task typically includes three scenarios: (1) learn a intent detection model with only $K$ examples for each intent~\citep{zhang2020discriminative,mehri2020dialoglue,efficient_intent};  (3) learn to identify both in-domain  and  out-of-scope queries with only $K$ examples for each intent~\citep{zhang2020discriminative,zhang2021pretrained, xia2021incremental}. (2) given a model trained on existing intents with all examples, learn to generalize the model to new intents with only $K$ examples for each new intent~\citep{xia2020composed,xia2020cg,xia2021pseudo}. 

In this work, we focus on the first scenario, and several methods have been proposed to tackle the challenge. Specifically, \newcite{zhang2020discriminative} proposes a  data augmentation schema, which pre-trains a model on annotated pairs from natural language inference (NLI) datasets and designs the nearest neighbor classification schema to adopt the transfer learning and classify user intents.  However, the training is expensive and hard to scale to tasks with hundreds of intents~\cite{liu2020basket}. \newcite{mehri2020example,efficient_intent} propose the task-adaptive training, which leverages models pre-trained from a few hundred million dialogues to tackle few-shot intent detection. It also includes an unsupervised mask language modeling loss on the target intent datasets and shows promising improvements.

Contrastive learning has shown superior performance on various domains, such as visual representation~\citep{he2020momentum,chen2020simple,radford2learning}, graph representation~\cite{qiu2020gcc,you2020graph}, and recommender systems~\cite{liu2021contrastive}. Moreover, recent works also adopt contrastive learning in natural language processing tasks~\citep{gunel2020supervised,liu2021fast,gao2021simcse}, which employs the contrastive learning to train the encoder. Specifically, \cite{gunel2020supervised} designs a supervised contrastive learning loss for fine-tuning data. \newcite{gao2021simcse}  designs a simple contrastive learning framework through dropout and it shows state-of-the-art performance on unsupervised and full-shot supervised semantic textual similarity tasks. \newcite{liu2021fast} designs self-supervised Mirror-BERT framework with two types of data augmentation: randomly erase or mask parts of the input texts; feature level augmentation through dropout.

Our work differs from them in several respects: Firstly, we specifically tackle the few-shot intent detection task rather than the general full-shot learning; Secondly, we design a schema and employ contrastive learning in both self-supervised pre-training and supervised fine-tuning stages. 

\section{CPFT Methodology}

We consider a few-shot intent detection task that handles $C$
user intents, where the task is to classify a user utterance $u$ into one of the $C$ classes.
We set balanced K-shot learning for each intent~\cite{zhang2020discriminative,efficient_intent}, i.e., each intent only includes $K$ examples in the training data. As such, there are in total $C\cdot K$ training examples.  

In the following section, we first describe the self-supervised contrastive pre-training for utterance understanding before introducing the supervised fine-tuning for few-shot intent detection.

\subsection{Self-supervised Pre-training}
We retrieve the feature representation $\mathbf{h}_i$ for the $i$-th user utterance through an encoder model, which in this paper is BERT~\cite{bert}, \textit{i.e.}, $\mathbf{h}_i=\text{BERT}(u_i)$. 
We implicitly learn the sentence-level utterance understanding and discriminate semantically similar utterances through the self-supervised contrastive learning method~\citep{wu2020clear,liu2021fast,gao2021simcse}:
\begin{equation}
    \label{eq:stage-1-unsupvised cl}
    \small
    \mathcal{L}_{\text{uns\_{cl}}}=-\frac{1}{N}\sum_{i=1}^{N}\log\frac{\exp(\text{sim}(\mathbf{h}_i,\bar{\mathbf{h}}_i)/\tau)}{\sum_{j=1}^{N}\exp(\text{sim}(\mathbf{h}_i,\bar{\mathbf{h}}_j)/\tau)},
\end{equation}
where $N$ is the number of sentences in a batch. $\tau$ is a temperature parameter that controls the penalty to negative samples. $sim(\mathbf{h}_i,\bar{\mathbf{h}_i})$ denotes the cosine similarity between two input vectors $\mathbf{h}_i$ and $\bar{\mathbf{h}_i}$.  $\bar{\mathbf{h}_i}$ represents the representation of sentence $\bar{u_i}$, where $\bar{u_i}$ is from the same sentence $u_i$ but few (10\%) tokens are randomly masked~\cite{bert}. Specifically,  we dynamically mask tokens during batch training~\citep{tod-bert}, i.e., a sentence has different masked positions across different training epochs, and we find it is beneficial to the utterance understanding. The sentence $u_i$ and  $\bar{u_i}$ are inputted together to a single encoder during the batch training~\citep{gao2021simcse}. 

Besides the sentence-level enhancement, we also add the mask language modeling loss~\cite{bert,tod-bert} to enhance the token-level utterance understanding:
\begin{equation}
    \label{eq:stage-1-mlm}
    \small
    \mathcal{L}_{\text{mlm}}=-\frac{1}{M}\sum_{m=1}^{M}\log P(x_m),
\end{equation}
where $P(x_m)$ denotes the predicted probability of a masked token $x_m$ over the total vocabulary, and $M$ is the number of masked tokens in each batch. 

Our total loss for each batch is $\mathcal{L}_{\text{stage1}}=\mathcal{L}_{\text{uns\_{cl}}}+\lambda\mathcal{L}_{\text{mlm}}$, where $\lambda$ is a weight hyper-parameter.

\subsection{Supervised Fine-tuning}
Through self-supervised learning in the first stage, the model efficiently utilizes many unlabeled user utterances. The model is given very limited examples in the second stage, such as 5 and 10 examples for each intent.  To better understanding user intents, especially when intents are similar to each other, we utilize a supervised contrastive learning method~\citep{gunel2020supervised} and train it together with an intent classification loss. We treat two utterances from the same class as a positive pair and the two utterances across different classes as a negative pair for contrastive learning. Unlike the previous work, the utterance and itself could also be a positive pair as we input them together to the single encoder. Their feature representations are different due to the dropout of BERT.  The corresponding loss is shown as the following:
\begin{equation}
 \small
     \mathcal{L}_{\text{s\_{cl}}}=-\frac{1}{T}\sum_{i=1}^{N}\sum_{j=1}^{N}\mathds{1}_{y_i=y_j} \log\frac{\exp(\text{sim}(\mathbf{h}_i,\mathbf{h}_j)/\tau)}{\sum_{n=1}^{N}\exp(\text{sim}(\mathbf{h}_i,\mathbf{h}_n)/\tau)},
\end{equation}
where $T$ is the number of pairs from the same classes in the batch. 

Next is the intent classification loss:
\begin{equation}
\small
    \mathcal{L}_{\text{intent}}=-\frac{1}{N}\sum_{j=1}^{C}\sum_{i=1}^{N}\log P(C_j|u_i),
\end{equation}
where $P(C_j|u_i)$ is the predicted probability of the $i$-th sentence to be the $j$-th intent class.

We jointly train the two losses together at each batch: $\mathcal{L}_{\text{stage2}}=\mathcal{L}_{\text{s\_{cl}}}+\lambda' \mathcal{L}_{\text{intent}}$, where $\lambda'$ is a weight hyper-parameter.

\begin{table}[]
\resizebox{\linewidth}{!}{
\centering
\begin{tabular}{lccc}
\hline
\textbf{Name} & \multicolumn{1}{l}{\textbf{\# Utterance}} & \multicolumn{1}{l}{\textbf{\# Intent}} & \multicolumn{1}{l}{\textbf{\# Domain}} \\ \hline
CLINC150~\citep{oos-intent}      & 18200                                     & 150                                    & 10                                     \\
BANKING77~\citep{efficient_intent}     & 10162                                     & 77                                     & 1                                      \\
HWU64~\citep{liu2019benchmarking}         & 10030                                     & 64                                     & 21                                     \\
TOP~\citep{gupta2018semantic}           & 35741                                     & 25                                     & 2                                      \\
SNIPS~\citep{coucke2018snips}         & 9888                                      & 5                                      & -                                      \\
ATIS~\citep{tur2010left}          & 4978                                      & 21                                     & -                           \\ \hline         
\end{tabular}}\caption{Data statistics for intent detection datasets.}\label{table:data-stats}
\end{table}


\begin{table*}[ht]
\centering
\resizebox{\linewidth}{!}{
\begin{tabular}{l|cc|cc|cc}
\hline
                                     & \multicolumn{2}{c|}{CLINC150} & \multicolumn{2}{c|}{BANKING77} & \multicolumn{2}{c}{HWU64} \\
Model                                & 5-shot        & 10-shot       & 5-shot        & 10-shot       & 5-shot      & 10-shot      \\ \hline
RoBERTa+Classifier~\citep{zhang2020discriminative}                   & 87.99         & 91.55         & 74.04         & 84.27         & 75.56       & 82.90         \\
USE~\citep{efficient_intent}                                  & 87.82         & 90.85         & 76.29         & 84.23         & 77.79       & 83.75        \\
CONVERT~\citep{efficient_intent}                              & 89.22         & 92.62         & 75.32         & 83.32         & 76.95       & 82.65        \\
USE+CONVERT~\citep{efficient_intent}                          & 90.49         & 93.26         & 77.75         & 85.19         & 80.01       & 85.83        \\
CONVBERT~\citep{mehri2020dialoglue}                             & -             & 92.10         & -             & 83.63         & -           & 83.77        \\
CONVBERT + MLM~\citep{mehri2020dialoglue}                       & -             & 92.75         & -             & 83.99         & -           & 84.52        \\
CONVBERT + Combined~\citep{mehri2020example} & -             & 93.97         & -             & 85.95         & -           & 86.28        \\
DNNC~\citep{zhang2020discriminative}                                 & 91.02         & 93.76         & 80.40         & 86.71         & 80.46       & 84.72        \\ \hdashline
CPFT                                & \textbf{92.34}         & \textbf{94.18}         & \textbf{80.86}         & \textbf{87.20}          & \textbf{82.03}       & \textbf{87.13}        \\ \hline
\end{tabular}}\caption{Testing accuracy ($\times 100\%$) on three datasets under 5-shot and 10-shot settings.}\label{table:main-results}
\end{table*}

\section{Experimental Settings}

\subsection{Datasets}

\paragraph{Pre-training Datasets}
We collected six public datasets consisting of different user intents. The dataset statistics are shown in Table  \ref{table:data-stats}.\footnote{\url{https://github.com/jianguoz/Few-Shot-Intent-Detection}} For fair comparisons, we exclude their test sets during the pre-training phase, which is different from previous work~\citep{mehri2020dialoglue,mehri2020example}, where they use the whole datasets. We also remove utterances with less than five tokens, and there are 80,782 training utterances in total. We conduct self-supervised pre-training on the collected utterances without using labels. 


\paragraph{Evaluation Datasets}
To better study the more challenging fine-grained few-shot intent detection problem and compare with recent state-of-the-art baselines,
we pick up three challenging intent detection datasets for evaluation, \textit{i.e.}, CLINC150~\citep{oos-intent}, BANKING77~\citep{efficient_intent} and HWU64~\citep{liu2019benchmarking}. CLINC150 contains 23,700 utterances across ten different domains, and there are in total 150 intents. BANKING77 contains 13,083 utterances with a single banking domain and 77 intents. HWU64 includes 25,716 utterances with 64 intents spanning 21 domains. We follow the setup of \newcite{mehri2020dialoglue}, where a small portion of the training set is separated as a validation set, and the test set is unchanged.  Following previous work, we repeat our few-shot learning model training five times and report the average accuracy. 

\subsection{Model Training and Baselines}

We utilize RoBERTa with {\tt base} configuration, \textit{i.e.}, {\tt roberta-base} as the BERT encoder. We pre-train the combined intent datasets without test sets in the contrastive pre-training stage for 15 epochs, where we set the batch size to 64, $\tau$ to 0.1, and $\lambda$ to 1.0. The pre-training phase takes around 2.5 hours on a single NVIDIA Tesla V100 GPU with 32GB memory. We fine-tune the model under 5-shot (5 training examples per intent) and 10-shot settings (10 training examples per intent). We set the batch size to 16, and do hyper-parameters search for $\tau \,{\in}\, \{0.1, 0.3, 0.5\}$  and $\lambda' \,{\in}\, \{0.01,0.03,0.05\}$; the fine-tuning takes five minutes for each run with 30 epochs. We apply label smoothing to the intent classification loss, following \newcite{zhang2020discriminative}. 

\paragraph{Baselines}
We compare with six strong models. 1, RoBERTa+Classifier~\citep{zhang2020discriminative}: it is a RoBERTa-based classification model. 2, USE~\newcite{yang2020multilingual}: it is the large multilingual model pre-trained on 16 languages. 3, CONVERT~\citep{efficient_intent}: it is an intent detection model with dual encoders, and the dual encoder models are pre-trained on 654 million (input, response) pairs from Reddit. 4, CONVEBERT~\citep{mehri2020dialoglue}: it fine-tunes BERT on a large open-domain dialogue corpus with 700 million conversations. 5, CONVEBERT+Comined~\citep{mehri2020example}: it is an intent detection model based on CONVEBERT, with example-driven training based on similarity matching and observers for transformer attentions. It also conducts task-adaptive self-supervised learning with mask language modeling (MLM) on the intent detection datasets. {\tt Combine} represents the best {\tt MLM+Example+Observers} setting in the referenced paper. 6, DNNC~\citep{zhang2020discriminative}: it is a discriminative nearest-neighbor model which finds the best-matched example from the training set through similarity matching. The model conducts data augmentation during training and boosts performance by pre-training on three natural language inference tasks.

\begin{table*}[]
\centering
\resizebox{\linewidth}{!}{
\begin{tabular}{l|cc|cc|cc}
\hline
                                                                  & \multicolumn{2}{c|}{CLINC150} & \multicolumn{2}{c|}{BANKING77} & \multicolumn{2}{c}{HWU64} \\
Model                                                             & 5-shot        & 10-shot       & 5-shot        & 10-shot       & 5-shot      & 10-shot      \\ \hline
CPFT                                                              & 92.34         & 94.18         & 80.86         & 87.20          & 82.03       & 87.13        \\ \hdashline
w/o Contrastive pre-training                                       & -4.15         & -2.63         & -4.11         & -2.37         & -6.01       & -4.17        \\
w/o Supervised contrastive learning                               & -0.56         & -0.32         & -2.06         & -0.88         & -1.14       & -0.27      \\
w/o Contrastive pre-training + w/o Supervised contrastive learning & -4.35         & -2.69         & -6.82         & -2.93         & -6.47       & -4.23        
  \\ \hline
\end{tabular}} \caption{Testing accuracy ($\times 100\%$) of CPFT with variants on three datasets under 5-shot and 10-shot settings.}\label{table:ablation-study}
\end{table*}

\section{Experimental Results}

 We show  the overall comparisons on three datasets in Table \ref{table:main-results}. The proposed CPFT method achieves the best performance across all datasets under both the 5-shot and 10-shot settings. Specifically,  CPFT outperforms DNNC by 1.32\% and 1.57\% on CLINC150 and HWU64 under the 5-shot setting, respectively. It also improves DNNC by 2.41\% on HWU64 under the 10-shot setting. 
 Our variances are also lower when compared with DNNC: Ours vs. DNNC: 0.39 vs. 0.57 and 0.18 vs. 0.42 on CLINC150; 0.20 vs. 0.88 and 0.48 vs. 0.21 on BANKING77; 0.51 vs. 1.00 and 0.25 vs. 0.38 on HWU64 under 5-shot and 10-shot settings, respectively.
 The improvements indicate that our proposed method has a better ability to discriminate semantically similar intents than the strong discriminate nearest-neighbor model with data augmentation.
Moreover, the DNNC training is expensive, as when training models on a single NVIDIA Tesla V100 GPU with 32GB memory, DNNC takes more than $3$ hours for 10-shot learning on CLINC150, and it needs to retrain the model for every new setting. CPFT only needs $2.5$ hours for one-time pre-training, and the  fine-tuning only takes five minutes for each new setting.  
Compared with CONVBERT+MLM, which does a self-supervised pre-training with MLM on the intent detection datasets, CPFT improves the performance by 1.43\%, 3.21\%, and 2.61\% on CLINC150, BANKING77, and HWU64 under 10-shot setting, respectively. CPFL also outperforms CONVBERT+Combined, which further adds examples-driven training and specific transformer attention design. We contribute the performance improvements to contrastive learning, which help the model discriminate semantically similar intents.

\section{Ablation Study and Analysis}
\paragraph{Is the schema with both stages necessary?}
We conduct ablation study to investigate the effects of self-supervised contrastive pre-training and supervised contrastive fine-tuning. Table \ref{table:ablation-study} shows the testing results of CPFT with model variants on three datasets. Experimental results indicate that both stages are necessary to achieve the best performance. The self-supervised contrastive pre-training on the first stage is essential as the performance drops significantly on all datasets. We hypothesize that contrastive pre-training on the intent datasets without using labels benefits the discrimination of semantically similar utterances. Additionally, the performance also drops if without supervised contrastive learning during the few-shot fine-tuning stage. Specifically, it drops by 2\%  on BANKING77 under the 5-shot setting; the reason is that BANKING77 is a single domain dataset with many similar intents, where supervised contrastive learning can explicitly discriminate semantically similar intents with very limited training examples. 
We also jointly train the first and second stages together, and compared with the proposed CPFT schema, we observe minimal improvements. The joint training is also costly as it requires retraining the model every time for new settings. 

\paragraph{Is contrastive pre-training beneficial to the target intent dataset? }
Additionally, we study whether contrastive pre-training can benefit the intent detection when excluding the target datasets. Specifically, we pre-train the model on the datasets except for the HWU64 dataset on the first stage and do few-shot learning on HWU64 during the second stage. Compared to the model without contrastive pre-training on the first stage, the performances are improved by 1.98\% and 1.21\% under 5-shot and 10-shot settings, respectively. The improvements indicate that the contrastive pre-training is helpful to transfer knowledge to new datasets. However, there are still performance drops compared to the contrastive pre-training, including the HWU64 dataset. Which shows that it is beneficial to include the target dataset during self-supervised contrastive learning. We leave whether self-supervised contrastive pre-training only on the target intent dataset benefits as a future study.

\paragraph{Is the training sensitive to hyper-parameters?}

We also study the effects of hyper-parameters of contrastive learning, \textit{i.e.}, the  temperature $\tau$ and weight $\lambda'$. We set $\tau \,{\in}\, \{0.05,0.1,0.3,0.5\}$ and $\lambda' \,{\in}\, \{0.01,0.03,0.05, 0.1\}$. In our primary experiments, we do not find $\tau$ has a notable influence during the self-supervised contrastive pre-training on the first stage.  Besides, we found that a batch size larger than 32 works well in the pre-training phase.
However, during the few-shot fine-tuning stage, when setting $\tau$ to a small value $0.05$, which heavily enforces the penalty to hard negative examples and $\lambda'$ to a large value $0.1$, which increases the weight of supervised contrastive learning loss, the performance drops significantly. In addition, the batch size  influences performance on this stage. Therefore, few-shot supervised contrastive loss is sensitive to hyper-parameters
when there are limited training examples. 
\section{Conclusion}
In this paper, we improve the performance of few-shot intent detection via contrastive pre-training and fine-tuning. It first conducts self-supervised contrastive pre-training on collected intent detection datasets without using any labels, where the model implicitly learns to separate fine-grained intents. Then it performs the few-shot fine-tuning based on the joint intent classification loss and supervised contrastive learning loss, where the supervised contrastive loss encourages the model to  distinguish intents explicitly. Experimental results on three challenging datasets show that our proposed method achieves state-of-the-art performance.


\section{Acknowledgements}
This work is supported in part by NSF under grants III-1763325, III-1909323,  III-2106758, and SaTC-1930941.  We thank the anonymous reviewers for their helpful and thoughtful comments.


\bibliographystyle{acl_natbib}
\bibliography{emnlp2020}


\end{document}


\maketitle

\setcounter{table}{5}
\setcounter{figure}{4}

\appendix


\section{Additional Notes on the Use of NLI}
\label{app:nli}

There are other tasks modeling relationships between sentences.
Paraphrase~\citep{paranmt} and semantic relatedness~\citep{semeval} tasks are such examples.
It is possible to automatically create large-scale paraphrase datasets by machine translation~\citep{ppdb}.
However, our task is not a paraphrasing task, and creating negative examples is crucial and non-trivial~\citep{selectional-preference}.
In contrast, as described above, the NLI setting comes with negative examples by nature.
The semantic relatedness (or textual similarity) task is considered as a coarse-grained task compared to NLI, as discussed in the previous work~\citep{jmt}, in that the task measures semantic or topical relatedness.
This is not ideal for the intent detection task, because we need to discriminate between topically similar utterances of different intents.
In summary, the NLI task well matches our objective, with access to large datasets.

\section{A Note on the Threshold Selection} \label{threshold-selection}
\label{app:threshold}
Our joint score ($\mathrm{Acc}_\mathrm{in} + R_\mathrm{oos}$) in Section~4.2 gives the same weight to the two metrics, $\mathrm{Acc}_\mathrm{in}$ and $R_\mathrm{oos}$, compared to other combined metrics like $(C_\mathrm{in}+C_\mathrm{oos})/(N_\mathrm{in}+N_\mathrm{oos})$.
Such a combined metric can put much more weight on the in-domain accuracy when $N_\mathrm{in}$ and $N_\mathrm{oos}$ are imbalanced; Table~2 shows such imbalance on the development set.
\citet{oos-intent} sacrificed the OOS recall a lot, and the trade-off with respect to the threshold selection was not discussed.

\section{Training Details} \label{training-details}

\begin{table*}[t]
\centering
\resizebox{1.0\linewidth}{!}{
\begin{tabular}{l|ccc|ccc}
\hline
           & \multicolumn{3}{c|}{\textbf{Single domain}}                 & \multicolumn{3}{c}{\textbf{All domains}}                 \\ \cline{2-7} 
           & \textbf{Learning rate} & \textbf{Epoch}    & \textbf{Run} & \textbf{Learning rate} & \textbf{Epoch} & \textbf{Run} \\ \hline
Classifier & \{1e-4, 2e-5, 5e-5\}   & \{15, 25, 35\}    & 10             & \{1e-4, 5e-5\}         & \{15, 25, 35\} & 5              \\
Emb-kNN    & \{1e-4, 2e-5, 3e-5\}   & \{7, 10, 20, 25, 35\} & 10             &   \{2e-5, 5e-5\}        &  \{3, 5, 7\}        & 5              \\
DNNC       & \{1e-5, 2e-5, 3e-5, 4e-5\}   & \{7, 10, 15\}     & 10             & \{2e-5, 5e-5\}         & \{3, 5, 7\}   & 5              \\ \hline
\end{tabular}}\caption{some hyper-parameter settings for a few models.}\label{table:hyper-paramter}
\end{table*}

\begin{table*}[t]
\centering
\begin{tabular}{l|cc}
\hline
           & \textbf{5-shot}                 & \textbf{10-shot}                \\ \hline
Classifier & \{bs: 50, ep: 25.0, lr: 5e-05\} & \{bs: 50, ep: 35.0, lr: 5e-05\} \\ \hline
Emb-kNN    & \{bs: 200, ep: 7.0, lr: 2e-05\} & \{bs: 200, ep: 5.0, lr: 2e-05\}  \\ \hline
DNNC       & \{bs: 900, ep: 7.0, lr: 2e-05\} &  \{bs: 1800, ep: 5.0, lr: 2e-05\} \\ \hline
\end{tabular}
\caption{Best hyper-parameter settings for a few models on the all-domain experiments, where {\tt bs} is batch size, {\tt ep} represents epochs, {\tt lr} is learning rate.}\label{table:hyper-paramter-best-all}
\end{table*}

\paragraph{Dataset preparation}
To use the CLINC150 dataset~\citep{oos-intent}\footnote{\url{https://github.com/clinc/oos-eval}.} in our ways, especially for the single-domain experiments, we provide a zip file {\tt data\_preprocess\_for\_emnlp2020.zip} accompanied with the paper submission.

\paragraph{General training}\label{appendix-general-training}
This section describes the details about the model training in Section~4.3.
For each component related to RoBERTa and SRoBERTa, we solely follow the two libraries, transformers and sentence-transformers, for the sake of easy reproduction of our experiments.\footnote{\url{https://github.com/huggingface/transformers} and \url{https://github.com/UKPLab/sentence-transformers}.}
The example code to train the NLI-style models is also available.\footnote{\url{https://github.com/huggingface/transformers/tree/master/examples/text-classification}.}
We use the {\tt roberta-base} configuration\footnote{\url{https://s3.amazonaws.com/models.huggingface.co/bert/roberta-base-config.json}.} for all the RoBERTa/SRoBERTa-based models in our experiments.
All the model parameters including the RoBERTa parameters are updated during all the fine-tuning processes, where we use the AdamW~\citep{adamw} optimizer with a weight decay coefficient of 0.01 for all the non-bias parameters.
We use a gradient clipping technique~\citep{clip} with a clipping value of 1.0, and also use a linear warmup learning-rate scheduling with a proportion of 0.1 with respect to the maximum number of training epochs. 

\paragraph{Pre-training on NLI tasks}\label{appendix-pre-training}
For the pre-training on NLI tasks, we fine-tune a {\tt roberta-base} model on three publicly available datasets, i.e., SNLI~\citep{snli}, MNLI~\citep{mnli}, and WNLI~\citep{wnli} from the GLUE benchmark~\citep{glue}.
The optimizer and gradient clipping follow the above configurations.
The number of training epochs is set to $4$; the batch size is set to $32$; the learning rate is set to $2e-5$.
We use a linear warmup learning-rate scheduling with a proportion of $0.06$ by following \citet{roberta}.
The evaluation results on the development sets are shown in Table~\ref{table-pretrain}, where the low accuracy of WNLI is mainly caused by the data size imbalance.
We note that these NLI scores are not comparable with existing NLI scores, because we converted the task to the binary classification task for our model transfer purpose.

\paragraph{Text pre-processing}
For all the RoBERTa-based models, we used the RoBERTa {\tt roberta-base}'s tokenizer provided in the transformers library.\footnote{\url{https://github.com/huggingface/transformers/blob/master/src/transformers/tokenization_roberta.py}.}
We did not perform any additional pre-processing in our experiments.

\paragraph{Hyper-parameter settings}\label{appendix-hyper-parameter}
Table~\ref{table:hyper-paramter} shows the hyper-parameters we tuned on the development sets in our RoBERTa-based experiments.
For a single-domain experiment, we take a hyper-parameter set and apply it to the ten different runs to select the threshold in Section~4.2 on the development set.
We then select the best hyper-parameter set along with the corresponding threshold, and finally apply the model and the threshold to the test set.
We follow the same process for the all-domain experiments, except that we run each experiment five times.
Table~\ref{table:hyper-paramter-best-all} and Table~\ref{table:hyper-paramter-best} summarize the hyper-parameter settings used for the evaluation on the test sets.
We note that each model was not very sensitive to the different hyper-parameter settings, as long as we have a large number of training iterations.

\begin{table}[]
\centering
\resizebox{\linewidth}{!}{
\begin{tabular}{l|lll}
\hline
Dataset                    & SNLI & WNLI & MNLI \\ \hline
Size of the development set & 9999          & 70            & 9814          \\
Accuracy                   & 94.5\%        & 41.4\%        & 92.1\%        \\ \hline
\end{tabular}}\caption{Development results on three NLI datasets.}
\label{table-pretrain}
\end{table}

\begin{table*}[t]
\centering
\resizebox{\linewidth}{!}{
\begin{tabular}{l|cccc}
\hline
           & \textbf{5-shot}                                      & \multicolumn{1}{c|}{\textbf{10-shot}}                 & \textbf{5-shot}                  & \textbf{10-shot}                 \\ \cline{2-5} 
           & \multicolumn{2}{c|}{\textbf{Banking}}                                                                                 & \multicolumn{2}{c}{\textbf{Credit cards}}                                    \\ \hline
Classifier & \multicolumn{1}{l}{\{bs: 15, ep: 25.0, lr: 5e-05\}}  & \multicolumn{1}{l|}{\{bs: 15, ep: 35.0, lr: 5e-05\}}  & \{bs: 15, ep: 15.0, lr: 5e-05\}  & \{bs: 15, ep: 25.0, lr: 5e-05\}  \\
Emb-kNN    & \multicolumn{1}{l}{\{bs: 200, ep: 35.0, lr: 1e-05\}} & \multicolumn{1}{l|}{\{bs: 200, ep: 25.0, lr: 2e-05\}} & \{bs: 100, ep: 20.0, lr: 1e-05\} & \{bs: 100, ep: 10.0, lr: 1e-05\} \\
DNNC       & \multicolumn{1}{l}{\{bs: 370, ep: 15.0, lr: 1e-05\}} & \multicolumn{1}{l|}{\{bs: 370, ep: 7.0, lr: 2e-05\}}  & \{bs: 370, ep: 15.0, lr: 2e-05\} & \{bs: 370, ep: 7.0, lr: 3e-05\}  \\ \hline
           & \multicolumn{2}{c}{\textbf{Work}}                                                                            & \multicolumn{2}{c}{\textbf{Travel}}                                 \\ \hline
Classifier & \{bs: 15, ep: 15.0, lr: 5e-05\}                      & \{bs: 15, ep: 15.0, lr: 5e-05\}                       & \{bs: 15, ep: 35.0, lr: 5e-05\}  & \{bs: 15, ep: 25.0, lr: 1e-04\}  \\
Emb-kNN    & \{bs: 100, ep: 20.0, lr: 1e-05\}                     & \{bs: 100, ep: 7.0, lr: 2e-05\}                       & \{bs: 100, ep: 35.0, lr: 3e-05\} & \{bs: 100, ep: 20.0, lr: 1e-05\} \\
DNNC       & \{bs: 370, ep: 7.0, lr: 3e-05\}                      & \{bs: 370, ep: 15.0, lr: 2e-05\}                      & \{bs: 370, ep: 7.0, lr: 2e-05\}  & \{bs: 370, ep: 7.0, lr: 2e-05\}  \\ \hline
\end{tabular}}\caption{Best hyper-parameter settings for a few models on the four single domains, where {\tt bs} is batch size, {\tt ep} represents epochs, {\tt lr} is learning rate.}\label{table:hyper-paramter-best}
\end{table*}

\section{Data Augmentation} \label{data-augmentation}

We describe the details about the classifier baselines with the data augmentation techniques in Section~4.3.

\paragraph{EDA}
Classifier-EDA uses the following four data augmentation techniques in \citet{eda}: synonym replacement, random insertion, random swap, and random deletion.
We follow the publicly available code.\footnote{\url{https://github.com/jasonwei20/eda_nlp}.}
For every training example, we empirically set one augmentation based on every technique.
We apply each technique separately to the original sentence and therefore every training example will have four augmentations.
The probability of a word in an utterance being edited is set to 0.1 for all the techniques.   

\paragraph{BT}
For classifier-BT, we use the English-German corpus in \citet{escape}, which is widely used in an annual competition for automatic post-editing research on IT-domain text~\citep{ape-2019}.
The corpus contains about 7.5 million translation pairs, and we follow the {\it base} configuration to train a transformer model~\citep{transformer} for each direction.
Based on the initial trial in our preliminary experiments to generate diverse examples, we decided to use a temperature sampling technique instead of a greedy or beam-search strategy.
More specifically, logit vectors during the machine translation process are multiplied by $\tau$ to distort the output distributions, where we set $\tau = 5.0$.
For each training example in the intent detection dataset, we first translate it into German and then translate it back to English.
We repeat this process to generate up to five unique examples, and use them to train the classifier model.
Table~\ref{tb:bt-examples} shows such examples, and we will release all the augmented examples for future research.

\begin{table*}[t]
  \begin{center}
{\small
    \begin{tabular}{l|l|l}
    
    Original utterance & Augmented example & Intent label \\ \hline
    can you block my chase account right away please & can you turn my chase account off directly & freeze account \\
    do a car payment from my savings account & with my saving account, you can pay a car payment account & pay bill \\
    when is my visa due & when is my visa to be paid & bill due \\ \hline

    \end{tabular}
}
    \caption{Examples used to train clasifier-BT.}
    \label{tb:bt-examples}
  \end{center}

\end{table*}

\section{More Results}
\label{extra-results}

\paragraph{Visualization} \label{appendix-vidualization}
Figure~\ref{fig:visulization-appendix} shows the same curves in Figure~3 along with the corresponding 10-shot results.
We can see that the 10-shot results also exhibit the same trend.
Figure~\ref{fig:tsne-appendix} shows more visualization results with respect to Figure~1.
Again, the 10-shot visualization shows the same trend.

Figure~\ref{fig:Conf-appendix} and Figure \ref{fig:Conf-appendix-all-domains} show 5-shot and 10-shot confidence levels on the test sets of the banking domain and all domains, respectively.
Both Classifier and Emb-kNN cannot perform well to distinguish the in-domain examples from the OOS examples, while DNNC has a clearer distinction between the two.

\paragraph{Faster inference}\label{appendix-DNNC-joint}
Figure~\ref{fig:joint_nli-appendix} shows the same curves in Figure~4 also for the 10-shot setting.
We can see the same trend with the 10-shot results.

\paragraph{Case studies}\label{Case Study}
Table~\ref{table:case-study} shows four DNNC prediction examples from the development set of the banking domain.
For the first example, the input utterance is correctly predicted with a high confidence score, and it has a similarly matched utterance to the input utterance;
for the second example, the input utterance is predicted incorrectly with a high confidence score, where the matched utterance is related to money but it has a slightly different meaning with the input utterance.
For the third example, the model gives a very low confidence score to predict an OOS user utterance as an in-domain intent; the last example is an incorrect case where the input utterance and the matched utterance have a topically similar meaning, resulting in a high confidence score for the wrong label, ``bill due.''
Based on these observations, it is an important direction to improve the model's robustness (even with the large-scale pre-trained models) towards such confusing cases.

\begin{figure*}[t]
	\begin{center}
    	\includegraphics[width=0.85\linewidth]{./images_appendix/Visualization_Metric_Appendix.png}
    \end{center}
\caption{5-shot and 10-shot development results on the banking domain. In this series of plots, a model with a higher area-under-the-curve is more robust.}
\label{fig:visulization-appendix}
\end{figure*}

\begin{figure*}[t]
	\begin{center}
    	\includegraphics[width=0.85\linewidth,height=0.9\textheight]{./images_appendix/TSNE_Appendix.png}
    \end{center}
\caption{5-shot and 10-shot tSNE visualizations on development set of the banking domain, where circles represent in-domain intent classes, and red stars represent out-of-scope intents.}
\label{fig:tsne-appendix}
\end{figure*}

\begin{figure*}[t]
	\begin{center}
    	\includegraphics[width=0.85\linewidth,keepaspectratio=true,height=0.95\textheight]{./images_appendix/Conf_Appendix_with_sbert.png}
    \end{center}
\caption{5-shot and 10-shot confidence levels on test set of the banking domain. Best viewed in color.}
\label{fig:Conf-appendix}
\end{figure*}

\begin{figure*}[t]
	\begin{center}
    	\includegraphics[width=0.85\linewidth]{./images_appendix/Conf_Appendix_All_domains.png}
    \end{center}
\caption{5-shot and 10-shot confidence levels on test set of all domains. Best viewed in color.}
\label{fig:Conf-appendix-all-domains}
\end{figure*}

\begin{figure*}[t]
	\begin{center}
    	\includegraphics[width=0.85\linewidth]{./images_appendix/Visualization_Joint_nli_Appendix.png}
    \end{center}
\caption{5-shot and 10-shot DNNC-joint development results on the banking domain, where the dash lines are DNNC results.}
\label{fig:joint_nli-appendix}
\end{figure*}

\begin{table*}[]
\centering
\resizebox{1.0\linewidth}{!}{
\begin{tabular}{ll}
\hline
\textbf{input utterance}    & transfer ten dollars from my wells fargo account to my bank of america account              \\
\textbf{matched utterance} & transfer \$10 from checking to savings                                                      \\
\textbf{label of the input utterance}      & transfer                                                                                    \\
\textbf{label of the matched utterance}      & transfer                                                                                    \\
\textbf{confidence score}             & 0.934                                                                                      \\ \hline
\textbf{input utterance}    & what transactions have i accrued buying dog food                                            \\
\textbf{matched utterance} & what have i spent on food recently                                                          \\
\textbf{label of the input utterance}      & transactions                                                                                \\
\textbf{label of the matched utterance}      & spending history                                                                           \\
\textbf{confidence score}             & 0.915                                                                                      \\ \hline
\textbf{input utterance}    & who has the best record in the nfl                                                          \\
\textbf{matched utterance} & do i have enough in my boa account for a new pair of skis                                   \\
\textbf{label of the input utterance}      & OOS                                                                                         \\
\textbf{label of the matched utterance}      & balance                                                                                     \\
\textbf{confidence score}             & 0.006                                                                                       \\ \hline
\textbf{input utterance}    & how long will it take me to pay off my card if i pay an extra \$50 a month over the minimum \\
\textbf{matched utterance} & how long do i have left to pay for my chase credit card                                     \\
\textbf{label of the input utterance}      & OOS                                                                                         \\
\textbf{label of the matched utterance}      & bill due                                                                                   \\
\textbf{confidence score}             & 0.945                                                                                      \\ \hline
\end{tabular}} \caption{Case studies on the development set of banking domain. The first two cases are in-domain examples from the banking domain, and the rest are OOS examples.}\label{table:case-study}
\end{table*}

\newpage
\bibliographystyle{acl_natbib}
\bibliography{emnlp2020}